\documentclass[pmlr,twocolumn,10pt]{jmlr}

\usepackage{booktabs} 
\usepackage{tikz}
\usepackage{url}
\usepackage{multicol}

\usepackage[draft,inline,nomargin,index]{fixme}
\fxsetup{theme=colorsig,mode=multiuser,inlineface=\itshape,envface=\itshape}
\FXRegisterAuthor{rp}{arp}{Richard}
\FXRegisterAuthor{sf}{asf}{Sorelle}

\usepackage{fancyvrb}

\usepackage{amsmath} 

\usepackage{multicol}

\newcommand{\mypara}[1]{\paragraph{#1}}

\usepackage[export]{adjustbox}

\usepackage[hyphenbreaks]{breakurl}

\jmlrvolume{81}
\jmlryear{2018}
\jmlrworkshop{Conference on Fairness, Accountability, and Transparency}

\title{Interpretable Active Learning\titletag{\thanks{This research was funded in part by the NSF under grants IIS-1633387, DMR-1709351, and DMR-1307801 and by the Arnold and Mabel Beckman Foundation.}\thanks{\url{https://github.com/rlphilli/InterpretableActiveLearning}}}}
\author{\Name{Richard L. Phillips} \Email{rlphillips@haverford.edu}\\
            \addr Haverford College
            \AND
            \Name{Kyu Hyun Chang} \Email{kyuhyun217@gmail.com}\\
            \addr Google Inc.\thanks{Work done while the author was a student at Haverford College.}
            \AND
            \Name{Sorelle A. Friedler} \Email{sorelle@cs.haverford.edu}\\
            \addr Haverford College
}

\begin{document}

\maketitle
 
\begin{abstract}
\looseness-1 Active learning has long been a topic of study in machine learning. However, as increasingly complex and opaque models have become standard practice, the process of active learning, too, has become more opaque. There has been little investigation into interpreting what specific trends and patterns an active learning strategy may be exploring. This work expands on the Local Interpretable Model-agnostic Explanations framework (LIME) to provide explanations for active learning recommendations. We demonstrate how LIME can be used to generate locally faithful explanations for an active learning strategy, and how these explanations can be used to understand how different models and datasets explore a problem space over time. 

In order to quantify the per-subgroup differences in how an active learning strategy queries spatial regions, we introduce a notion of uncertainty bias (based on disparate impact) to measure the discrepancy in the confidence for a model's predictions between one subgroup and another.  Using the uncertainty bias measure, we show that our query explanations accurately reflect the subgroup focus of the active learning queries, allowing for an interpretable explanation of what is being learned as points with similar sources of uncertainty have their uncertainty bias resolved.  We demonstrate that this technique can be applied to track uncertainty bias over user-defined clusters or automatically generated clusters based on the source of uncertainty.
\end{abstract}

\begin{keywords}
Interpretability; Active learning.
\end{keywords}

\section{Introduction}
The importance of interpretability and explainability of machine-learned decisions has recently been an area of active interest, with the EU even declaring what has been called a ``right to an explanation" \cite{EUexplanationright}.  In traditional machine learning contexts, the focus of interpretability has been two-fold, first on the receiver of the decision (``why was I rejected for this job?") and second on the model creator (``why is my model giving these answers?").

Here, we extend this interest in interpretability to active learning, a domain in which the explanation is additionally of interest to the labeler (``why am I being asked these questions and why is it worth it to answer?").  Since active learning is generally applied in scenarios such as drug discovery where it is expensive (whether in terms of time or money) to label a query, the labeler in these contexts is often a domain expert in their own right (e.g., a chemist).  Given this, a query explanation can serve as a way to both justify an expensive request and allow the domain expert to give feedback to the model.

\subsection{Results}
We demonstrate how active learning choices can be made more interpretable to non-experts.  Using per-query explanations of uncertainty, we develop a system that allows experts to choose whether to label a query.  This allows experts to incorporate domain knowledge and their own interests into the labeling process.  In addition, we introduce a quantified notion of \emph{uncertainty bias}, the idea that an algorithm may be less certain about its decisions on some data clusters than others.  In the context of decision-making about people, this may mean that some protected groups (e.g., race or gender) may receive less favorable decisions due to risk aversion \cite{EUexplanationright}.  In the context of active learning, this means that these groups are more likely to be targeted for exploratory queries in order to improve the model.  We combine this idea with the explanations generated per query to describe the groups most targeted by uncertainty bias.  More broadly, these techniques allow us to make active learning interpretable to expert labelers, so that queries and query batches can be explained and the uncertainty bias can be tracked via interpretable clusters.

\section{Related Work}
\mypara{Active Learning}
Active learning has a long history detailed in a comprehensive survey by \citet{settles.tr09}.  Our work will focus on explaining query uncertainty.  Uncertainty querying for active learning was first proposed by \citet{Gale}. Since then, it has become perhaps the most common strategy for active learning and  several strategies for quantifying uncertainty have been developed \cite{settles.tr09}. Strategies used to quantify uncertainty for actively learning multi-class classification problems include selecting the sample with the minimum maximum-class probability, selecting the sample with the minimum difference in probabilities between the two most probable classes, and choosing the sample with maximal label entropy.  All three of the above strategies are equivalent for the binary classification tasks we will focus on in this paper \cite{settles.tr09}.

Related to our focus on the added impact of a domain expert on an active learning system, \cite{BaldridgePalmer2009} focus on evaluating the strength of active versus passive learning with expert versus novice labelers.  They found that the domain expert was able to take advantage of the more effective active learning setting, while the novice labeler did not provide the expected increase in performance from active learning.  While that work focused on the impact of an expert on the labeling process, we will focus on the impact of an expert on the \emph{choice of query} to label.  Perhaps the closest work to what we propose here is the work of \cite{Glass2006Explaining}, which focuses on explaining preference learning results to users of a scheduling system.  However, that work aims to explain the way the preferences have been updated in a way that is specific to the model and domain problem, while we focus on explaining the choice of query using a general model in any application area. 

\mypara{Interpretability}
Another area of direct importance to this paper is interpretable machine learning.  Recent work on interpretability has included both local explanations about an individual's decision \cite{LIME} and global explanations about the model's actions overall, including interpretable techniques in clustering \cite{interpretableClustering}, integer programming \cite{SLIMrecidivism}, rule lists \cite{WangRudin2015Falling}, and methods for understanding deep nets \cite{Zeiler2014Visualizing, Le2013Building} in addition to historical work on decision trees \cite{Quinlan1993C4.5} and random forests \cite{breiman2001random}.

\mypara{LIME}
We will build specifically on a method for creating local explanations introduced in \cite{LIME}. Local Interpretable Model-Agnostic Explanations (LIME) is a framework for generating locally faithful explanations for an otherwise opaque machine learning model. LIME does this by taking a given point and perturbing it to generate many hypothetical points in the neighborhood of a query point and then training an interpretable model on this hypothetical neighborhood. More specifically, LIME generates an explanation for the prediction $f(x)$ for a selected point $x$ and a given a global classifier $f$. To generate this explanation, LIME requires a number of samples $N$, a similarity kernel $\pi$, a discretizer $d$, and a number of attributes $K$ to use for the explanation. LIME begins by generating $N$ vectors with Gaussian random noise scaled to the mean variance of the dataset as a whole. The discretization method $d$ transforms points from a continuous representation to a categorical one by reducing the values into bins. A local regressor is fit to the generated points and the global model's predicted class probabilities. This model is used to select features by iteratively re-fitting and greedily adding them to the model until $K$ features are found. \footnote{Alternatively, the features with the highest weights could be used or, if utilizing LASSO, the LASSO regularization path could be used.}  The similarity kernel $\pi$ is used to upweight points near the query point.Finally, LIME returns the chosen $K$ features and their weights in the final local model as the explanation for $f(x)$.

\begin{figure*}[!htb]
\centering
\centering
  \begin{tabular}{@{}ccc@{}}

\includegraphics[width=.3\textwidth]
{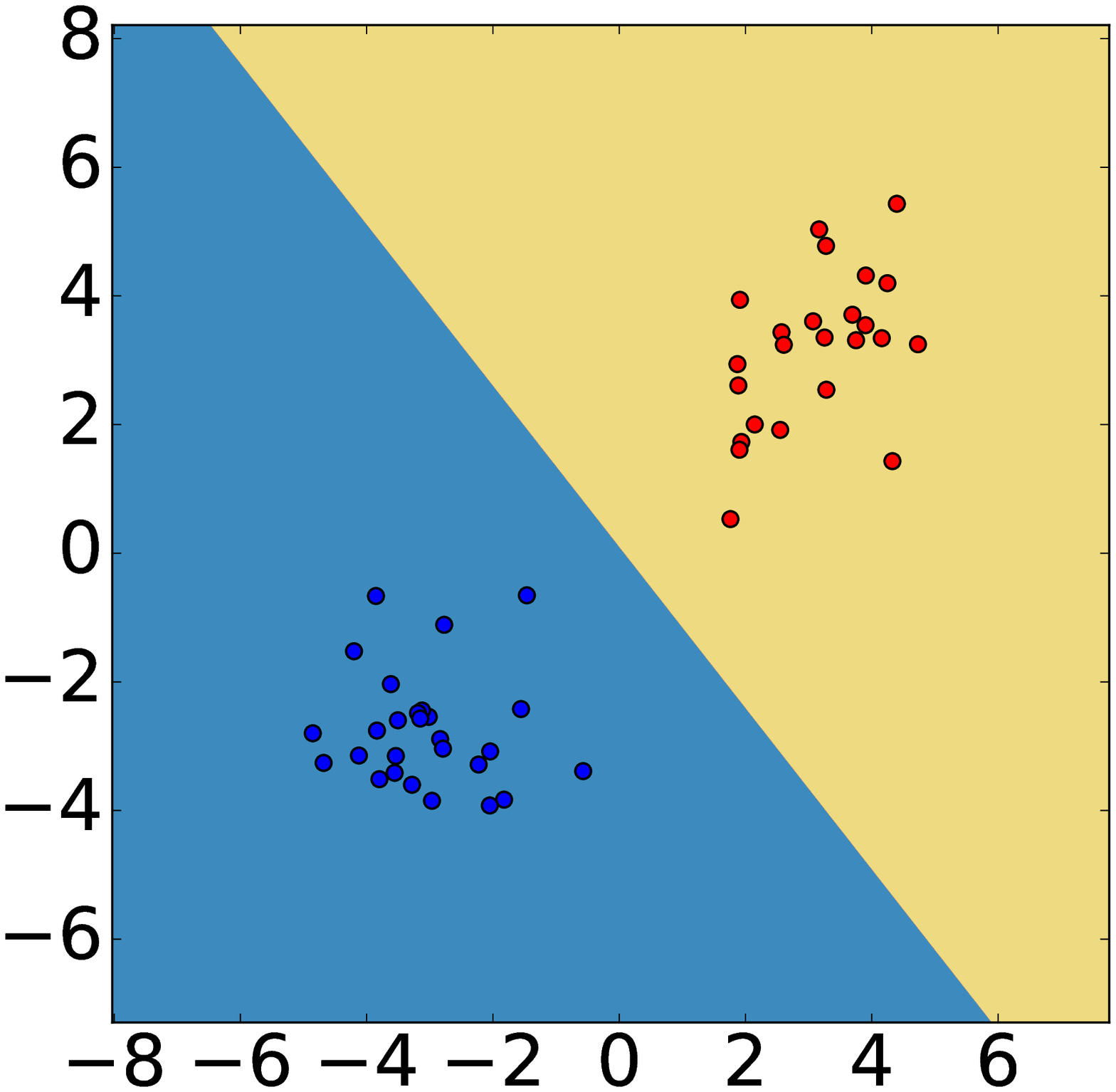} &
\includegraphics[width=.3\textwidth]{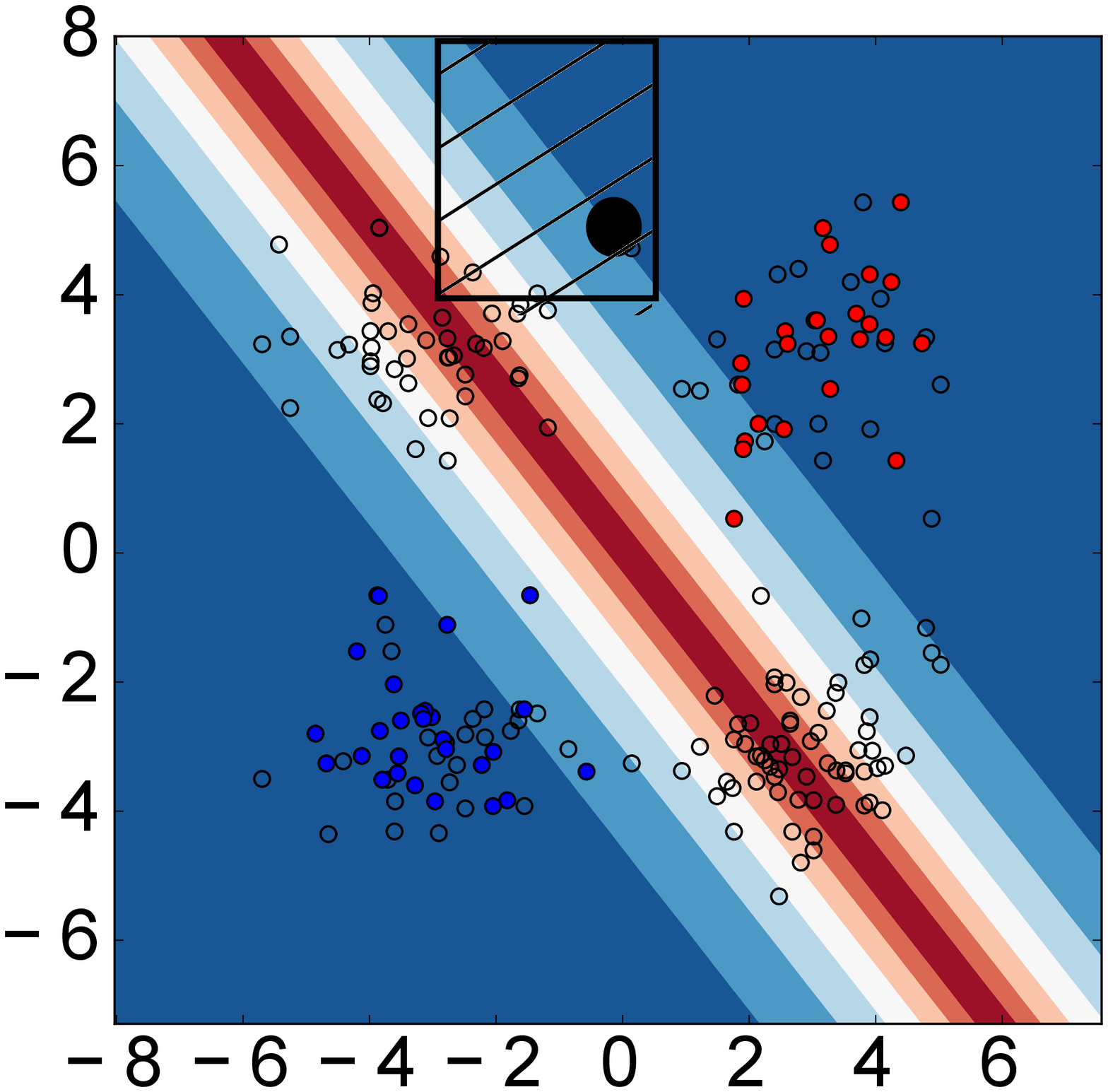} &
\includegraphics[width=.3\textwidth]{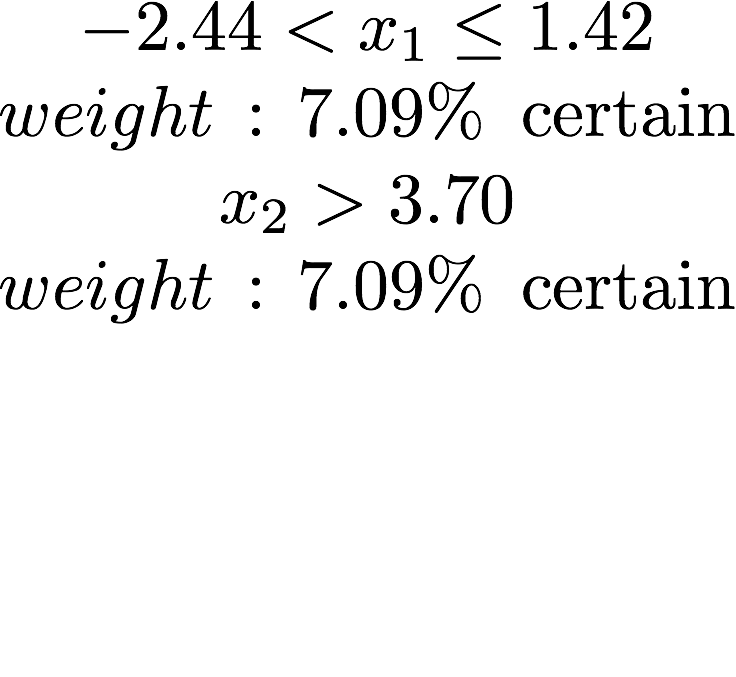}
 \end{tabular}
\caption{Left: Labeled starting pool. Center: Certainty over the problem space. The black points represent unlabeled points in the pool, blue represents regions of certainty, and red represents uncertainty. The large black point is the example query. Right: The LIME uncertainty explanation for the model's prediction of $95$\% uncertainty for the black point at $(0,5)$.   The associated uncertainty region is shown dashed in the center figure.}
\label{fig:toy_uncertainty}
\label{fig:toy_expla}
\end{figure*}

\section{Explaining Active Learning Queries}
\label{sec:basic_method}

We propose an application of LIME to explain the uncertainty for an individual points. For a query point $x$, the classifier $f$ has some certainty function $c$. Using LIME, we generate the local sample of $N$ points and fit a local regressor to these points (weighted by the kernel $\pi$) and the certainties provided by $c$. Using the standard LIME methodology, we generate explanations for a point's uncertainty rather than it's decision between two classes. 
To explain this technique further, we'll describe the method on a toy dataset.

\paragraph{Toy dataset} 
Four Gaussian distributions with unit variance are created and centered at $(-3,-3)$, $(3,-3)$, $(3,3)$, and $(-3,3)$ and labeled so that
the first two represent one class and the second another. Fifty initial points are randomly selected from the Gaussian at $(-3,-3)$ and $(3,3)$ to be labeled. The points have been purposefully drawn so as to label none of the points from the Gaussians centered in the second and fourth quadrants. An initial logistic regression model, $f$, is trained on the 50 labeled points.

Based on the resulting model of the probability distribution, the certainty scores $c$ across the problem space are mapped. The labeled points, decision boundary, and certainty scores can be seen in Figure \ref{fig:toy_uncertainty}.  LIME allows us to take an individual point and explain a local regressor around the uncertainty for said point, generating our resulting \emph{uncertainty explanation}.  Each uncertainty explanation is composed of one inequality \emph{constraint} for each of the included $K$ features, generating an associated \emph{uncertainty region} bounded by those constraints.

\subsection{Experiments}
 
In addition to our toy dataset described above, we will perform experiments on the ProPublica dataset for recidivism prediction \cite{propublica} as well as the Dark Reactions Project dataset of chemical reactions for synthesis prediction \cite{drpNature}.

\mypara{ProPublica dataset}
The ProPublica dataset includes attributes describing the sex, age, race, juvenile felony and misdemeanor counts, number of adult priors, charge degree (felony or misdemeanor), and charge description for 6172 people arrested in Broward County, Florida, along with a boolean value indicating whether they were rearrested within two years of the original arrest date.  We followed the cleaning steps recommended by \cite{propublica}\footnote{\url{https://github.com/propublica/compas-analysis}} and used a logistic regression model trained on an initial pool of 400 randomly selected points. 

\mypara{Dark Reactions Project dataset}
The Dark Reactions Project dataset includes 6114 hydrothermal synthesis reactions and 274 attributes describing chemical properties that might predict the associated boolean classification indicating whether the experiment successfully created a crystalline product or not. To predict this outcome, we used AdaBoost with 200 decision stump weak learners. The certainty function thus estimates the probability of each class through a weighted average of the fraction of training samples within each leaf of the decision stumps. 

\mypara{LIME Setup}
Our experiments use LIME \cite{LIME} to explain continuous (regressor) predictions.\footnote{\url{https://github.com/marcotcr/lime/}}\textsuperscript{,}\footnote{ \url{https://github.com/datascienceinc/lime}}   
We apply this to our active learning selection criterion. Ridge regression is the local `interpretable' model to estimate feature importance, as that is the default in the LIME library. This can be changed without altering our general framework. Continuous features were split into at most 8 bins by greedily maximizing information gain to make the splits. A value $K$ for the length of explanations to use is also selected; $K=2$ for the toy and ProPublica datasets and $K=6$ for the Dark Reactions dataset. 

On each of these datasets and initially trained models, we performed active learning using uncertainty sampling to maximize the class probability. Using the technique described above, we were able to explain each query before it was labeled. 

\mypara{ProPublica data set}
On the ProPublica data set, an example query instance with attributes $sex = Male$, $age=34$, $race=African-American$, $age\_category=25 - 45$, $juvenile\_felony\_count=2$, $juvenile\_misdemeanor\_count=1$, $priors\_count = 21$, and $charge\_degree $ $=$ $ felony$ was found to have certainty $0.88$.  The generated uncertainty explanation was:
~\\~\\
\begin{tabular}{rll}
$priors\_count > 20$ & \texttt{weight: 0.3} & and \\
$sex\_Male > 0.5$ & \texttt{weight: -0.04} &
\end{tabular}
~\\~\\
\noindent This means that this query point is fairly certain, primarily due to the high number of prior convictions. 
However, the fact that this person is a male slightly reduces the classifier's certainty for this point (indicated by the negative weight), perhaps due to the higher variance in recidivism for men than for women.  

\mypara{Dark Reactions data set}
An example Dark Reactions uncertain query is the reaction with $0.1648$ grams of Oxovanadium(2+) sulfate, $1.2995$ grams of Selenium dioxide, and $0.1166$ grams of 1,4-dimethylpiperazine heated at a temperature of $110$ Celsius, with a pH of 4, for 24 hours with a slow cooling process.  The resulting uncertainty explanation was: 
~\\~\\
\begin{tabular}{rl}
$1.76 < PaulingElectronegGeom <= 1.89$ & and \\
$1.76 < PaulingElectronegMean <= 1.91$ & and \\
$orgminimalprojectionradiusMax <= 2.84$ & and \\
$slowCool = True$ & and \\
$orghbdamsaccGeomAvg <= 0.99$ & and \\
$numberInorg = 2$
\end{tabular}
~\\~\\
\noindent  Normally, a reaction with lopsided ratios of the inorganic reactants would be likely to fail, however the small organic molecule projection radius indicates a common family of organic templates used successfully in the Dark Reaction Project, and the slow cool attribute increases the likelihood of a reaction succeeding, thus explaining the uncertainty in this query.  Note that there is an overlap in the uncertainty ranges for \texttt{PaulingElectronegMean} and \texttt{PaulingElectronegGeom}. This makes sense as these features encode almost the same information: the former is the mean and the latter is the geometric mean.

\section{Identifying Uncertainty Bias}
In order to provide an interpretable explanation of how active learning is proceeding as a whole, and not just an explanation for a single query as provided in the previous section, we will be interested in tracking how subgroups of the data are collectively queried based on their uncertainty.  The developed uncertainty explanations are made of a set of constraints that bound an \emph{uncertainty region} that can be useful for grouping contained points based on identical sources of uncertainty.  We expect points explored in an  uncertainty region to increase the certainty we have about other points in that region (that have the same sources of uncertainty).\footnote{In this work, LIME defined boundaries that together form a region are mutually exclusive sets, but this is not required for the technique.}

In situations where some instance populations are smaller (e.g., minority groups in a dataset where each item is a person) or where the initial training data distribution is skewed, the active learner may prefer queries that are disproportionately drawn from a single uncertainty region (or population group). For example, in our toy example above, we saw that the upper left quadrant is underrepresented in the labeled dataset. The points in this region have higher uncertainty and were more likely to be targeted for active learning queries.  In order to understand both what and how an active learning method is learning and whether there is skew in the uncertainty regions (subgroups) targeted to be labeled, we will quantify the \emph{uncertainty bias} of a subgroup.

\begin{definition}[Uncertainty bias]
Given a dataset $D = (\mathbb{X}, U)$ with $d$-dimensional feature vector $\mathbb{X}$ and corresponding (discrete) uncertainty labels $U$ and its disjoint set $R$ of uncertainty regions (groups), let $X_r = \{ x \in \mathbb{X} | x \in r, r \in R \}$ be the items of the dataset within a region $r \in R$.  $U$ takes values $+$ (certain) and $-$ (uncertain).  The \emph{uncertainty bias} with respect to region $r \in R$ is defined to be:
\[ 1 - \frac{Pr(U = + | x \in r)}{Pr(U = + | x \in R \setminus r)} \]
\end{definition}
\noindent Note that this uncertainty definition is the same as $1 - DI$ where $DI$ is the disparate impact value \cite{2015_kdd_disparate_impact} applied where the region of focus is the protected class and the positive value is a label of $+$.  For the purposes of this work, we consider any point with certainty greater than or equal to the median over our pool to be certain $(U = +)$. Discrete classes of uncertainty are used instead of continuous conditional probabilities to follow the existing literature defining disparate impact in law and machine learning \cite{Barocas16DisparateImpact, 2015_kdd_disparate_impact}.  Although we use discrete classes to calculate this definition, we use the actual uncertainty value for creating the explanations, as described in Section \ref{sec:basic_method}.

The goal of this definition is to identify subgroups that differ from the majority in terms of their uncertainty in the classifier, i.e., it is assumed that a good outcome would be for all subgroups to have the same amount of uncertainty.  In the context where subgroups are protected groups (e.g., race or gender), this goal translates to an aim to identify subgroups that might be  disproportionately queried.  Such subgroups might be subject to harms due to these queries (e.g., extra law enforcement scrutiny) or due to the uncertainty about their subgroup (e.g., risk aversion) \cite{EUexplanationright}, and this definition coupled with the associated interpretability methods aims to help identify and explain the sources of uncertainty for these groups.

\subsection{Tracking Uncertainty Bias Over Time}

There are many possible uncertainty regions that could be analyzed for uncertainty bias.  Often, a domain expert may have a specific set of features that they are interested in exploring.  In this set of experiments, we'll assume such feature sets are given (in Section \ref{sec:auto_class} we'll consider how to perform this analysis when they are not).  Using active learning with uncertainty sampling, these experiments track both the per-region count of chosen queries and the uncertainty bias per uncertainty region with each queried point.

\begin{figure}[h]
\centering
\includegraphics[width=.191739130435\textwidth]{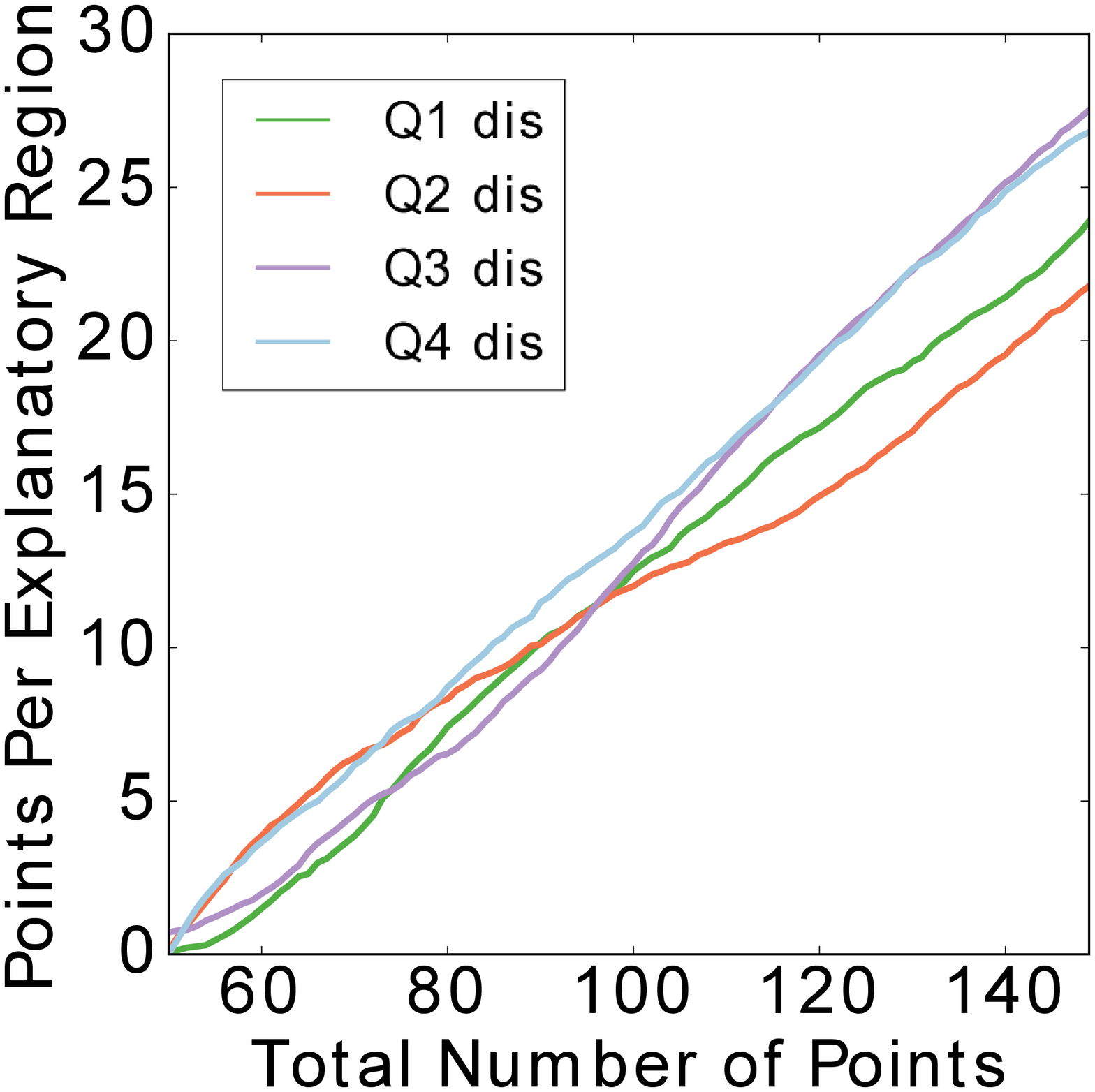}~~
\raisebox{0.132\height}{\includegraphics[width=.21\textwidth]{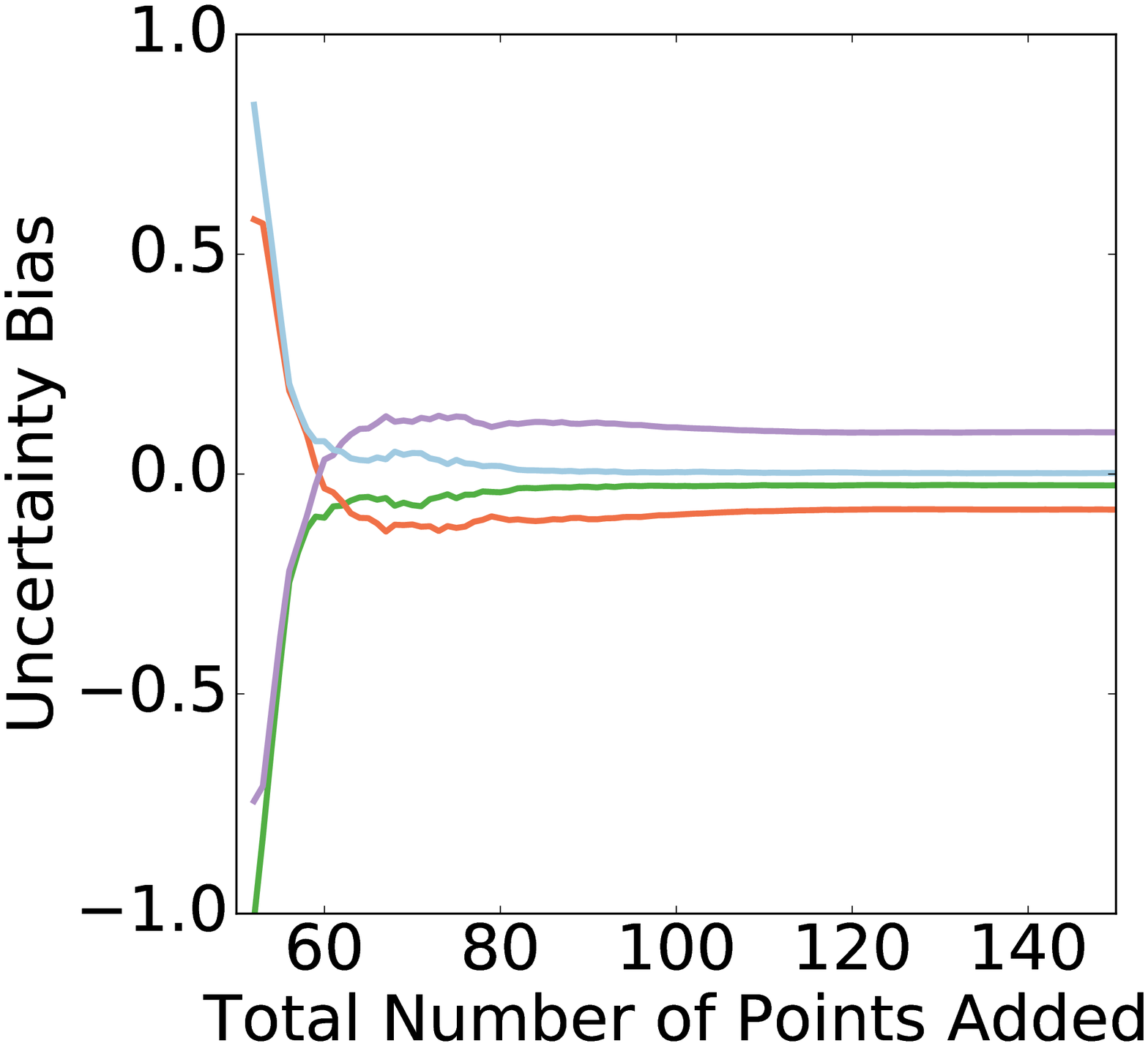}}
\caption{Left: Average counts of points taken from each quadrant during 50 active learning runs on the toy dataset. Right: Uncertainty bias per quadrant over 50 active learning runs.}
\label{fig:toy_illustrate}
\end{figure}

\mypara{Toy dataset}
In order to test the technique, we first perform these experiments on our toy dataset. Given the randomness of the choice of initial training set, results are averaged over 50 runs. From our knowledge of the toy dataset, we know that each quadrant has a centered distribution and so it makes sense to track the uncertainty bias using each of the four quadrants as an uncertainty region. We would expect to see the uncertainty bias of Quadrants 2 and 4, which begin with no points in the initial training set, start high and then fall. Indeed, the results on the toy dataset (see Figure \ref{fig:toy_illustrate}) are consistent with these expectations. 

\begin{figure*}[h]
\begin{center}
\includegraphics[width=.45\textwidth]{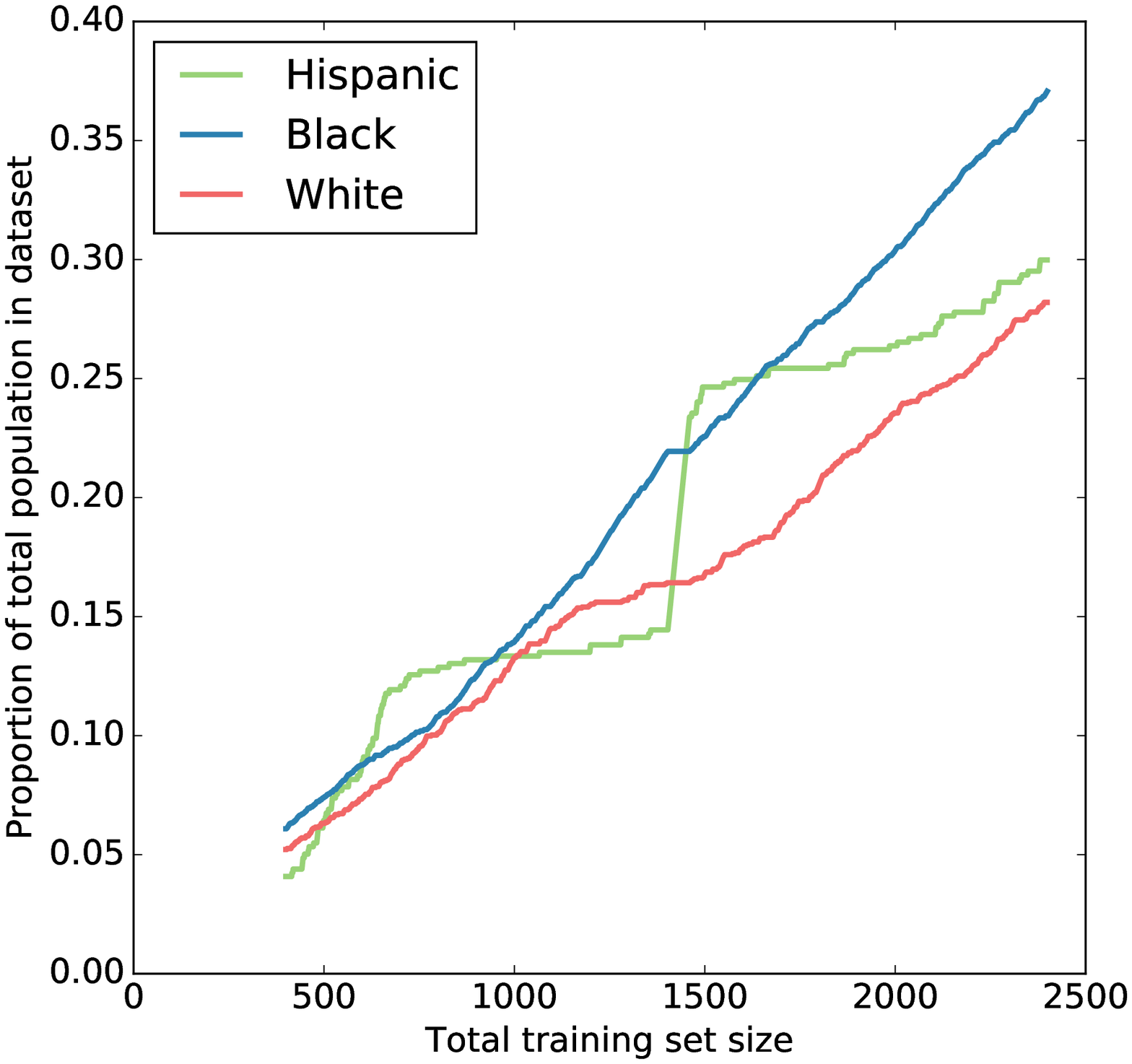}~~
\includegraphics[width=.45\textwidth]{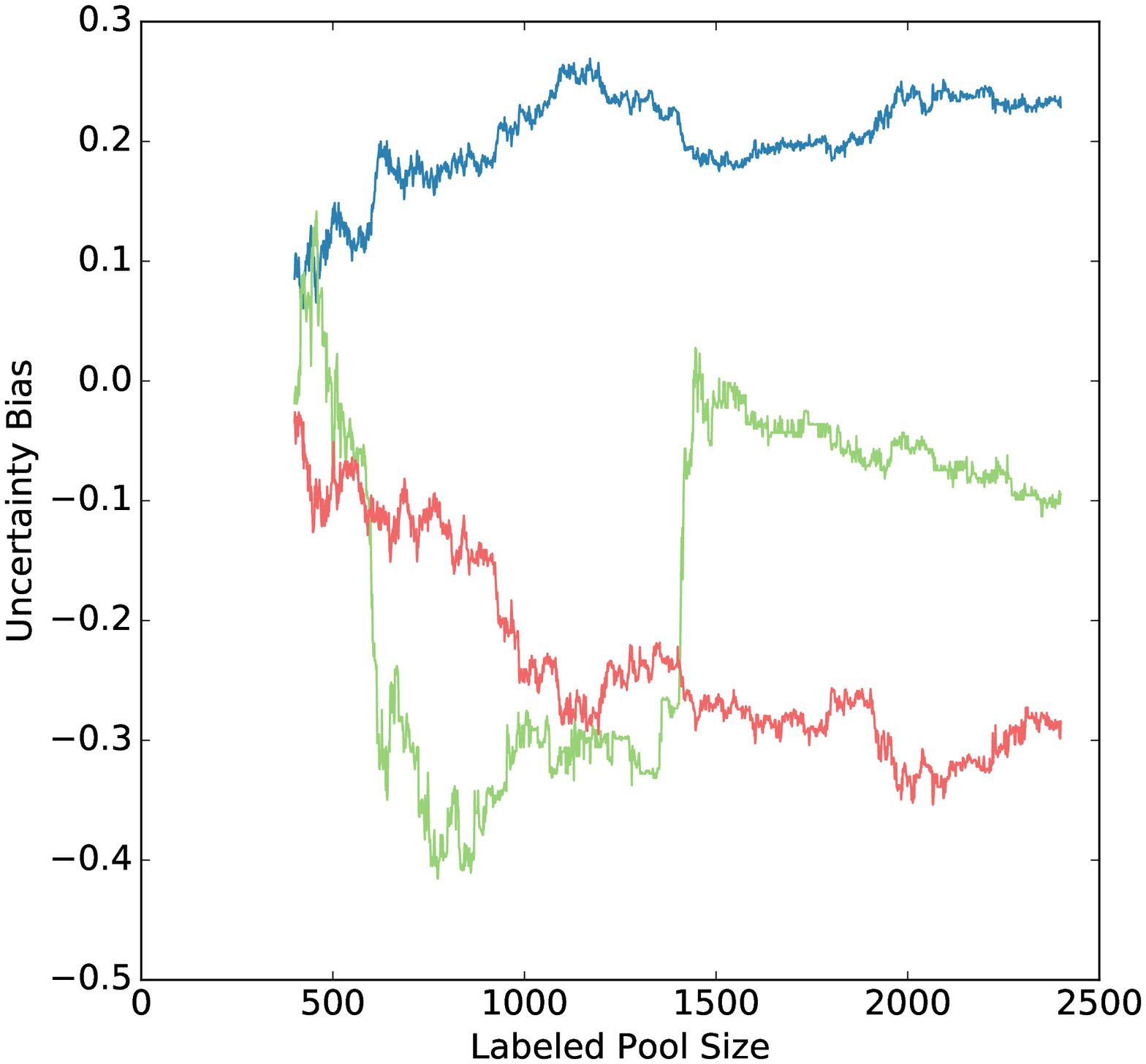}
\caption{Left: The proportion of the three subpopulations conditioned on race that were either in the initial set or have since been queried by the active learning algorithm for the ProPublica dataset experiment. Right: The corresponding uncertainty bias for each subpopulation conditioned on race as active learning queries are made.}
\label{fig:race_comp}
\end{center}
\end{figure*}

\begin{figure}[h]
\begin{center}
\includegraphics[scale=0.68]{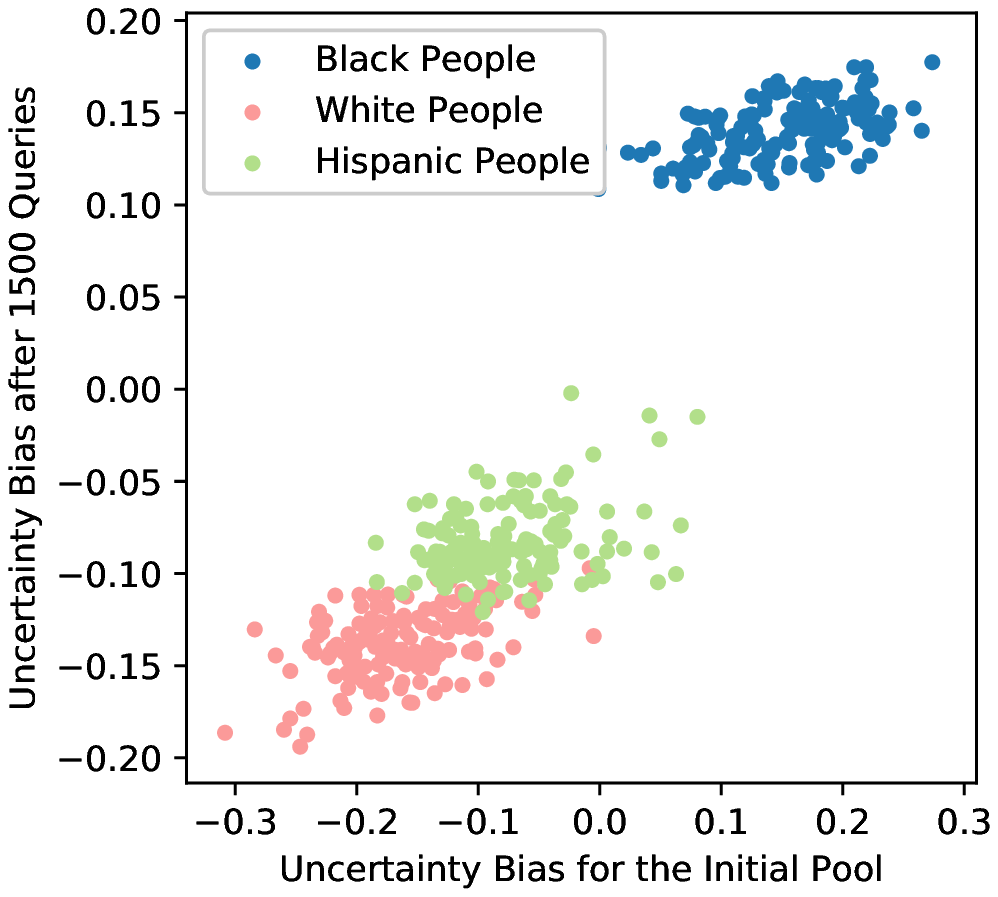}
\caption{Initial uncertainty biases for the three racial subgroups versus uncertainty biases after 1500 actively learned points have been added to the labeled dataset for 150 runs. Note the clear positive trend for each subgroup. This trend shows that initial uncertainty bias is highly influential on the latter uncertainty bias.}
\label{fig:race_bias_corr}
\end{center}
\end{figure}

\mypara{ProPublica dataset}
We next consider the ProPublica recidivism prediction dataset.  Since one of the main questions about this data is its potential for negative racial impact when used in classification, a natural question to consider is whether there is uncertainty bias when regions group people by race.  Using the same experimental setup, we tested the model for uncertainty bias based on race with each point added to our pool. The results can be found in Figure \ref{fig:race_comp}. It is evident that, from the very beginning, there is a notable disparity in our model's ability to make confident predictions between the different racial categories. While active learning seems to resolve most of the difference in uncertainty bias between the people labeled `White' and `Hispanic,' Black people arrested in Broward County were subject to considerable uncertainty bias by our logistic regression model even after 2000 more points were actively queried. This is true even considering that there are more Black people

\subsection{The Impact of the Initial Pool on Long-Term Uncertainty Bias}
One of the major lasting issues with many methods of active learning is sampling bias. Under the assumption that the initial labeled pool was selected randomly, sampling bias refers to the trend that this initially representative sample will diverge further from the true distribution during active learning as we overemphasize boundary-hugging points. In reality, initial labeled pools are generally not randomly sampled, but have been selected by humans based on some nonrandom criteria. With this in mind, we consider whether or not an active learning algorithm can help compensate for disparities in an initial training set over time to remedy disparities that may exist.

The race-label subclasses on the Propublica data seem to imply that active learning does not resolve human-introduced disparities in the initial dataset. However, we would like to explore what the effect of the initial dataset is. That is, does an initially highly biased dataset tend to stay biased under an active learning regime more than a lightly biased dataset? Will remedying biases in an initial dataset encourage parity between subgroups over the long term?

\paragraph{Propublica Data}
To explore the effect that the uncertainty bias of the initial pool has on the long-term uncertainty bias between groups, we generated 150 different starting pools. Each pool had 400 labeled, randomly selected candidates, just as in the previous ProPublica experiment. For each of these starting pools, we recorded the initial uncertainty bias for each racial subgroup and the uncertainty biases as we actively labeled 1500 queries one at a time for each starting pool. The results are shown in Figure \ref{fig:race_bias_corr}.\footnote{For black people in the dataset, the coefficient of determination was $r^2=  0.2586$, for white people $r^2= 0.2390$, and for hispanic people $r^2= 0.1163$. Overall $r^2= 0.8808$, but we do not necessarily expect a linear model to fit the data.} The x-axis values represent the initial starting uncertainty for every racial subgroup in each run. The y-axis values represent the uncertainty biases at the end of each trial.

Within each racial subgroup, there is a clear positive trend between the initial uncertainty bias and the resulting uncertainty bias after the 1500 additional points are actively queried. This supports our hypothesis that the initial relative uncertainties have lasting effects on the uncertainty bias of the model over the labeled set. That the resulting dataset remains proportionally biased after it has more than quadrupled in size, from 400 labeled points to 1900, implies strong and long-lasting effects for the initial pool. This is true even though there are more Black defendants in the dataset than white defendants (3,175 vs. 2,103 for the whole dataset). This implies that researchers should take greater care when manually assembling initial labeled datasets. That is, researchers should seek to rectify biases in an initial dataset \textit{before} applying active learning. Otherwise, these biases will be perpetuated even as the active learning algorithm progresses. With that said, however, the limited size of the ProPublica dataset limits our ability to explore how long lasting the effects of the initial pool can actually be. Figure \ref{fig:race_bias_corr} is also notable in that it clearly illustrates the racial hierarchy that is reestablished with each initial random pool and active learning run. 

Now that we have illustrated the utility of uncertainty bias in measuring the relative progress of subgroups in a dataset over a period of active learning, we can aim it at evaluating a novel batching strategy centered around uncertainty explanations.

\section{Batching Based on Uncertainty}
In practice, performing queries one-by-one can be highly inefficient. For instance, conducting biological or chemical experiments is time-consuming, but multiple experiments can be conducted at the same time. So in these settings it is more efficient to get a batch of queries, conduct several experiments at once, and provide a batch of labels (see, e.g, the drug discovery batch active learning process employed in \cite{NIPS2001_2097}). There are also interpretability considerations for one-by-one queries. That is, single queries may make it difficult for researchers to balance long term goals. For instance, single recommendations provide little indication of what long term patterns will be explored. Single queries may also hinder researchers from forming hypotheses before exploring problem sub-spaces, which may lead to retrospective justifications or possible inefficiencies in experimental design. \\
\indent Instead of simply providing a batch of queries that are expected to improve the underlying model the most, our goal is to provide a batch that also has the same uncertainty source, so that a single explanation of uncertainty can be provided to the oracle for the whole batch.

\subsection{Creating Interpretable Batches}
In order to create batches with an explainable uncertainty source, we consider each uncertainty region as a candidate for batch selection within that region instead of from the entire pool.  The batch selected from an uncertainty region is likely to have lower variance within the batch compared to a batch selected from the entire pool. Although this reduction in variance may lead to a performance loss, the benefits to this batch selection strategy will be in its interpretability.   

The number of unlabeled instances that are within an uncertainty region varies in size for different queries. It is possible that the number of unlabeled instances is smaller than the desired batch size. In order to cope with this, a bigger batch consisting of several smaller batches can be used.  When the number of unlabeled instances is larger than the batch size, a query selection strategy should be used where instances within the uncertainty region serve as the pool.

Once a batch is chosen from an uncertainty region, all queries share the same uncertainty explanation.  In addition, the specific items in the batch can be explained to the labeler by identifying the subset of features most useful to the labeler for interpretability and displaying the range or variation in those features within the batch.

\subsection{Experiments}
\begin{figure}[htbp]
\begin{center}
\includegraphics[width=.5\textwidth]{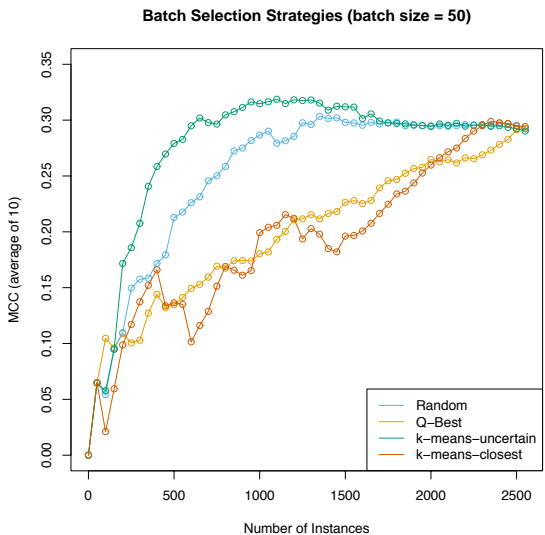}
\caption{A comparison of batch selection strategies for choosing queries within uncertainty regions on the Dark Reactions data set.}
\label{fig:drp_batch}
\end{center}
\end{figure}

On the Dark Reactions data set, we evaluated 4 batch selection strategies within positive uncertainty regions: Q-best, where the highest uncertainty instances were chosen (a random selection) and $k$-means, both where the instance closest to the center was chosen and where the most uncertain instance in the cluster was included in the batch.  Evaluating the resulting quality of the model using the Matthews Correlation Coefficient (a measure that takes into account true and false positives and negatives) and a batch size of 50, $k$-means (with $k=50$) when the most uncertain instance per cluster was the best batch selection strategy for this data (see Figure \ref{fig:drp_batch}). 

Using this batch selection strategy and the interpretability method above to describe the uncertainty region, we were able to generate explanations of uncertainty associated with each chosen batch (with a batch size of 20).  A subset of the features were chosen corresponding to the reactants and other experimental parameters that would be understandable to a human labeler (a chemist).  The variety of these features within the chosen batch was then described.

\begin{figure}[htbp]
\begin{center}
\includegraphics[width=.5\textwidth]{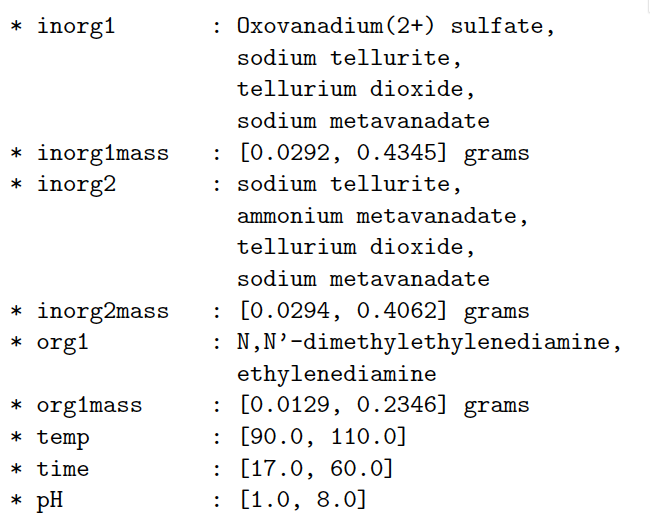}
\caption{An example explanation for a query batch from the Dark Reactions chemical data set.}
\label{fig:drp_batch_expl}
\end{center}
\end{figure}

One example explanation is shown in Figure \ref{fig:drp_batch_expl}. This specific query takes two similar organic templating molecules (listed next to `org1') and varies the manual experimental parameters for reactions that have high value as active learning candidates. While  every combination of the listed parameters may not be included in the batch, it is clear that this batch is exploring a specific organic templating structure over a varied swath of inorganic compounds, times, and pHs. Although there is likely some overlap in the information that these reactions will provide, they will make a model able to make much stronger predictions for reactions with similar organic templating molecules and may quickly create a more developed local hypothesis space relative to standard batching techniques as we vary the axes on which uncertainty is explored. 

\section{Interpretable Uncertainty Clusters}
\label{sec:auto_class}

In order to look at the idea of grouping things based on their sources of uncertainty in a more flexible and precise way, we will generate clusters based on point uncertainty explanations.  After we generate LIME explanations of uncertainty, we generate a new encoding for our points based on the explanation constraints and the weights that LIME assigns to them. That is, we construct a one-hot encoding of the categorical constraints that LIME provides and then replace the values where a LIME explanation constraint is present with the weight LIME provides for that feature. This means that for a set of possible uncertainty labels $U$, all of our original data points have an equivalent point in our encoding space ${\rm I\!R}^{|U|}$. Utilizing the weights instead of a standard one-hot encoding allows us to discriminate between different magnitudes of a feature's contribution to the local certainty estimate as well as between LIME explanations that indicate certainty and those that indicate uncertainty. As either direction will make a feature influential on the total uncertainty, LIME will provide explanations comprised of both. This is an improvement over the batch technique described in the previous section, which only considered positive sources of uncertainty that had the same combinations of uncertain traits. 

To automatically create groups for tracking the principle patterns explored during active learning, \textit{k}-means clustering is used to cluster the samples' explanations and weights. The objective of \textit{k}-means (using Lloyd's algorithm) is thus to minimize the pairwise squared deviations for all of the points in each cluster: $\sum_{i}^{k} \sum_{d \in {U}} \sum_{x,y \in C_i} \| x_d - y_d \| ^2$.  Each cluster centroid is then used to keep track of the principle sources of uncertainty for that cluster. The number of clusters, \textit{k} is chosen by trying a wide range of potential values and finding the value that maximizes the proportion of points that share their top uncertainty constraints with their respective cluster centroids. As this ratio will likely continue to trend upwards as \textit{k} grows, \textit{k} is simply increased until adding another \textit{k} will not improve this proportion over some small threshold. It is possible to largely capture all of the uncertainty labels for a pool within a relatively small number of clusters, greatly simplifying the task of tracking what regions of uncertainty are explored. The goal of this method is not to choose perfect interpretable clusters, but to demonstrate that creating interpretable groupings based on uncertainty is possible. We fully expect that there may be other and better methods of doing this.

\subsection{Experiments}

We will automatically generate interpretable uncertainty clusters for the ProPublica and Dark Reactions datasets and consider the uncertainty bias for each of these clusters.  For both experiments, we will train a model with an initial training set and iteratively add labeled points using active learning. We will track both the number of points queried from each cluster and the per-cluster uncertainty biases.

\begin{figure*}[h]
\centering
\includegraphics[width=.5\textwidth]{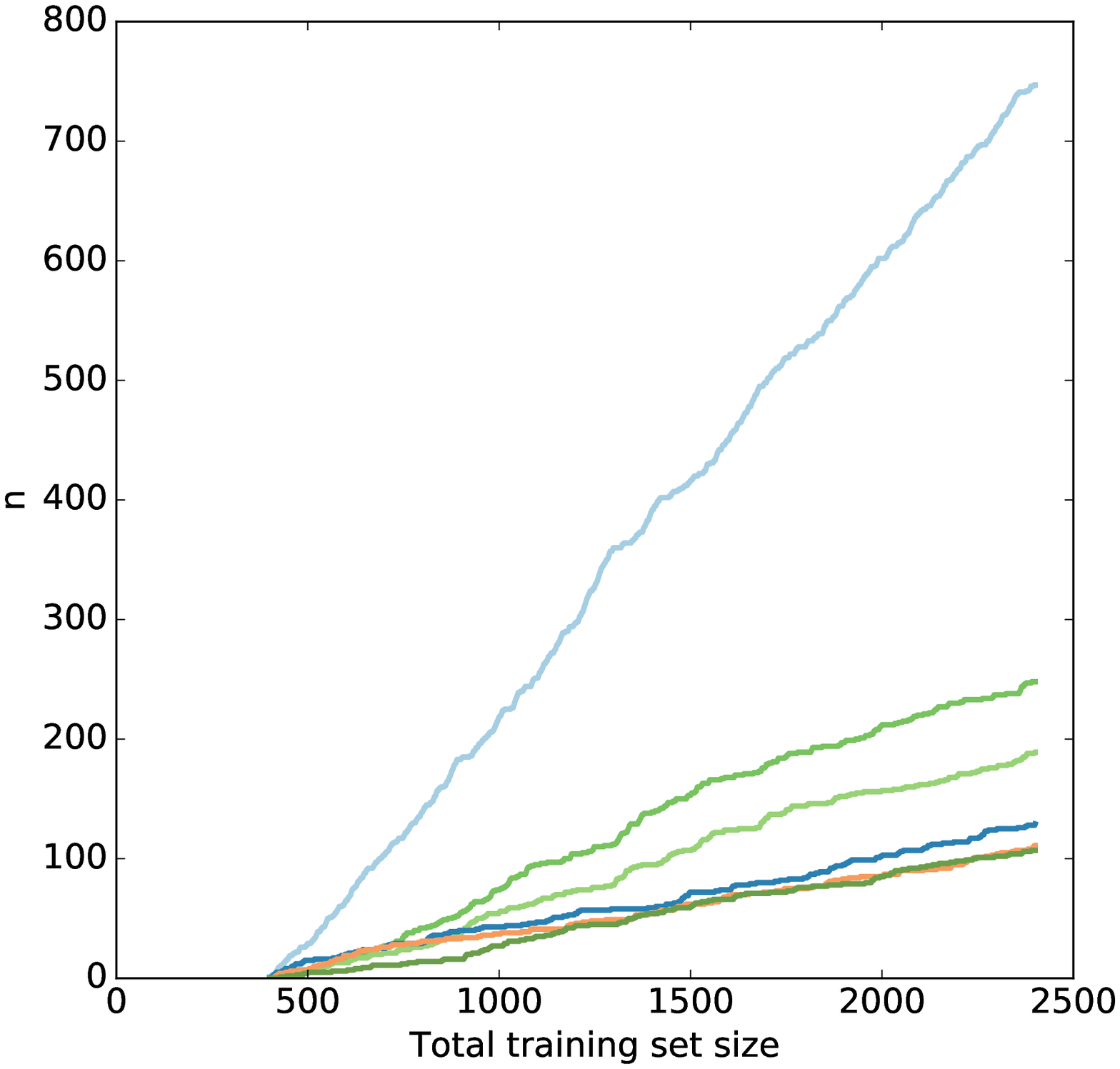}~~
\includegraphics[width=.5\textwidth]{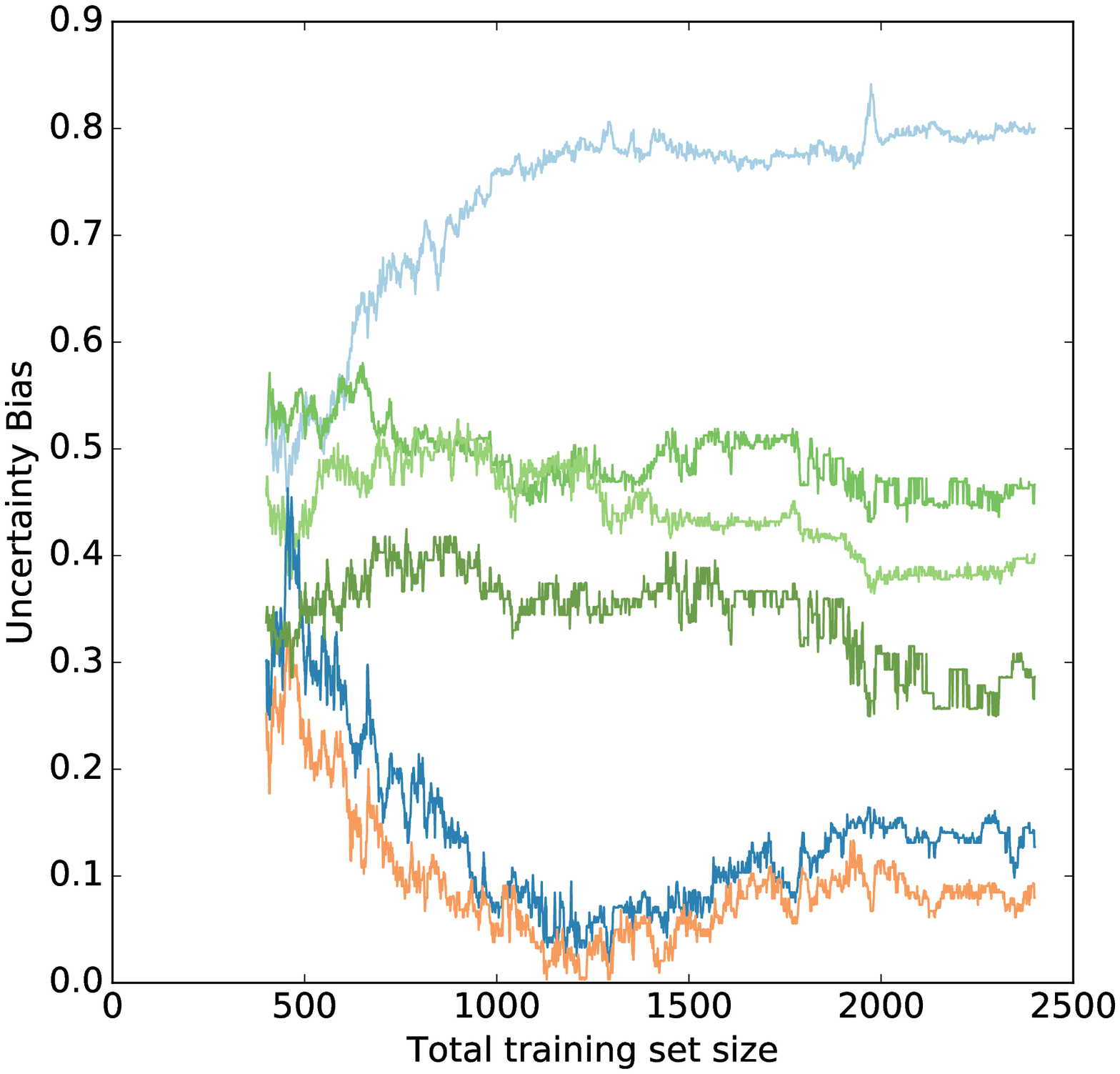}
\includegraphics[width=.5\textwidth]{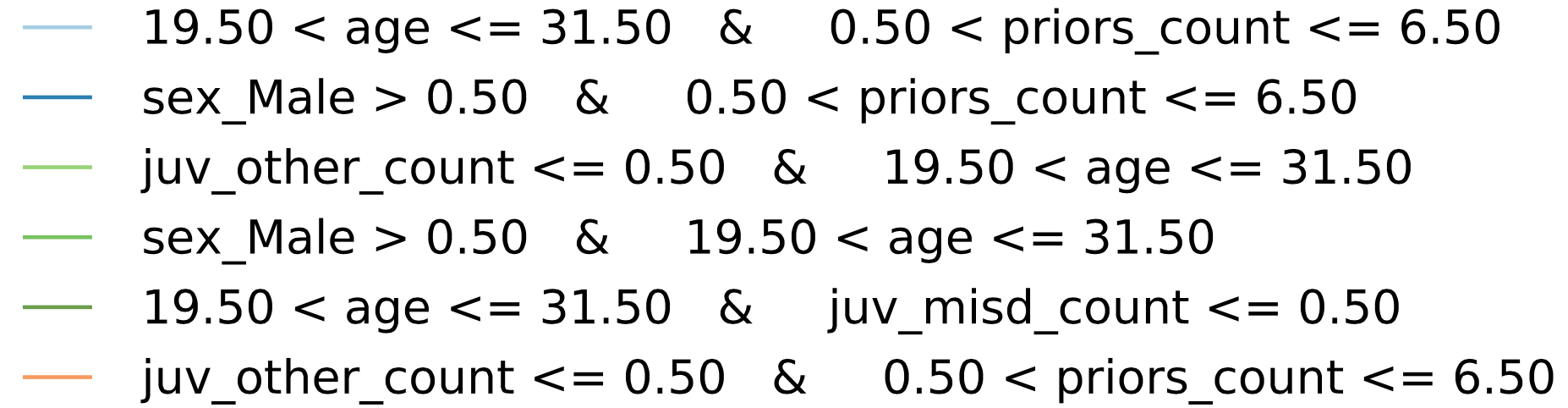}
\caption{Left: Counts of labeled points from each of the top six labeled clusters as queries are made. Right: The corresponding per-cluster uncertainty bias values.}
\label{fig:propublica_clusters}
\end{figure*}

\mypara{ProPublica data}
Each data point was given an uncertainty explanation with two constraints $(K=2)$ and the uncertainty labels and weights were clustered with $k=40$ clusters. Eighty-three percent of the points in the pool were in clusters with centroids that matched their own uncertainty constraints and 100\% shared at least one uncertainty label with their cluster's centroid. This shows that cluster centroids are indicative of the points in their respective clusters and that we can rely on the uncertainty labels for the centroid to understand a cluster's major sources of uncertainty. The resulting uncertainty bias charts per-cluster can be found in Figure \ref{fig:propublica_clusters}.

The first two clusters, covering people in their 20s and men with a few priors, have high uncertainty. One possible reason for this is that a high number of priors might make the chance of recidivism more certain and the age range and gender are likely very common in the dataset and naturally have high variance. Five of the six clusters all display overall downward trends, showing that uncertainty sampling is effectively able to reduce the uncertainty on these clusters.  It also provides support for the validity of our explanatory uncertainty labels, as the frequency with which clusters are queried correlates with the resolved uncertainty that a model has for those clusters.

\begin{figure*}[h]
\centering
\includegraphics[width=5in]{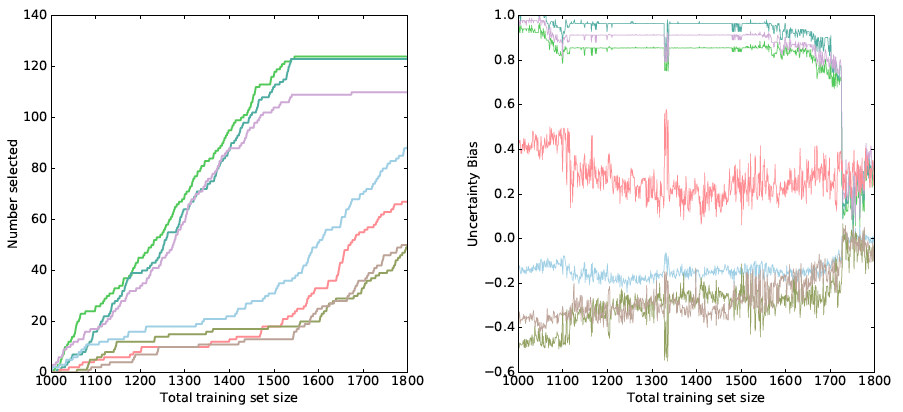}\\
\includegraphics[width=5in]{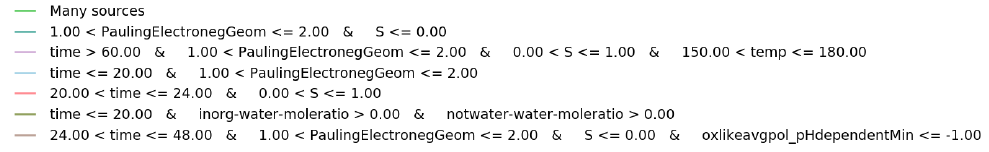}
\caption{Left: Counts labeled from each of the seven automatically-generated clusters for the Dark Reactions dataset. Right: Corresponding uncertainty bias values for the top six most queried clusters for the logistic regression model trained at each step.}
\label{fig:DRP_stuff}
\end{figure*}

\mypara{Dark Reactions dataset}
The Dark Reactions data set was a challenge because there are many more features, and yet fewer points, than the the ProPublica dataset. We considered $k = 7$ clusters and the associated uncertainty bias values. To produce more specific class labels than the ProPublica experiment, $K=10$ attribute constraints were used to explain the certainty of each point rather than $K=2$.  As $K$ is much higher for each explanation, consider the difference between the certainty for an individual constraint and the average certainty over all $K$ constraints for a given uncertainty explanation.  If the absolute value of that difference is at least 2\% (i.e., that specific constraint contributes more than average to the overall certainty), then it is included in the explanation.  By allowing a domain expert to adjust this 2\% cutoff parameter to be higher or lower, we can adjust towards more precise explanations of uncertainty.  Explanations with no attribute above the 2\% threshold receive the explanation ``many sources."  The results can be found in Figure \ref{fig:DRP_stuff}.

\looseness-1 Given that there are 274 attributes in the Dark Reactions dataset, it is notable that most of the curves do have prominent sources of uncertainty.  However most explored cluster has no attributes that rise above the 2\% threshold, indicating that all $K=10$ attributes contribute similarly to the uncertainty of that cluster, possibly indicating that it is highly varied.  Three of the clusters are explored together, and looking at the explanations we can see that they have similar sources of uncertainty.  Overall, the observed labels fit the intuition a domain expert might have for these reactions.  For instance, middle temperature, high time reactions (represented by the purple line) begin with a higher uncertainty bias than short reactions (represented by both the brown-green and light blue lines). Reactions that take less than 20 hours generally do not have the time to form crystalline products of the size required for this type of chemistry. Thus, these have high certainty because most will fail. In contrast, middle temperature, long time reactions have high potential to be successful, although many are not, causing the uncertainty.

\section{Discussion}
One of the issues with the strategy outlined in this work, as well as other strategies for explaining individual predictions in model interpretability, is that individual features or combinations of said features are not always interpretable themselves. This was a significant challenge with the Dark Reactions Project dataset. For instance, anyone who has taken an undergraduate chemistry class would be able to deduce that $O\_mols$ refers to the amount of oxygen in a reaction. However, even experienced chemists lack good intuition for some descriptors. For example, a descriptor in the DRP that conveys the range for the geometric means of atomic radii weighted by stoichiometry aggregated across compounds in the organic role in a reaction. There are many descriptors like this and, although a trained chemist can certainly explain what this descriptor means and how it is calculated, there is little good intuition for what it would mean if a high value of that descriptor were a source of uncertainty. Combinations of features help to make up for this by providing more detail and increasing the likelihood that there are several interpretable features from which to explain uncertainty. Even if there were not, however, several individually uninterpretable or unintuitive features could be brought together and they all, for instance, might inform one another as they all reference similar physical attributes. For chemists, at least, this issue of uninterpretable features is partially remedied by the batch recommendations, which paint a fuller picture of what is being explored. 

\section{Conclusion}
This work has demonstrated an application of LIME to explain active learning queries. We also defined a quantitative measure of uncertainty bias. We first demonstrated how we can track the exploration of groups of points with common uncertainty and confirmed that uncertainty is being resolved using the uncertainty bias measure. We then demonstrated on more complex, real-world datasets how regions of uncertainty can be generated automatically to create meaningful groups to track during learning. 

\bibliography{bibliography} 

\end{document}